\documentclass{article}

\usepackage[final]{corl_2026} 
\usepackage{graphicx}
\usepackage{amsmath}
\usepackage{multirow}
\usepackage{booktabs}
\usepackage{amssymb}
\usepackage{caption}
\usepackage[table]{xcolor}

\title{HumanX: Toward Agile and Generalizable Humanoid Interaction Skills from Human Videos}

\author{Yinhuai Wang\textsuperscript{1,2}$^*$, Qihan Zhao\textsuperscript{1}$^*$, Yuen Fui Lau\textsuperscript{1}$^*$, Runyi Yu\textsuperscript{1}, Hok Wai Tsui\textsuperscript{1}\\
\textbf{Qifeng Chen\textsuperscript{1}$^{\dagger}$, Jingbo Wang\textsuperscript{2}$^{\dagger}$, Jiangmiao Pang\textsuperscript{2}, Ping Tan\textsuperscript{1}$^{\dagger}$}\\
\textsuperscript{1}The Hong Kong University of Science and Technology\quad \textsuperscript{2}Shanghai AI Laboratory
\\
$^*$Equal contribution\quad $^{\dagger}$Corresponding authors
\\
\textbf{\textcolor[RGB]{117,189,66}{https://wyhuai.github.io/human-x/}}
}

\begin{document}

\maketitle
\thispagestyle{empty}
\pagestyle{empty}
\begin{center}
    \centering
    \captionsetup{type=figure}
    \includegraphics[width=1.0\textwidth]{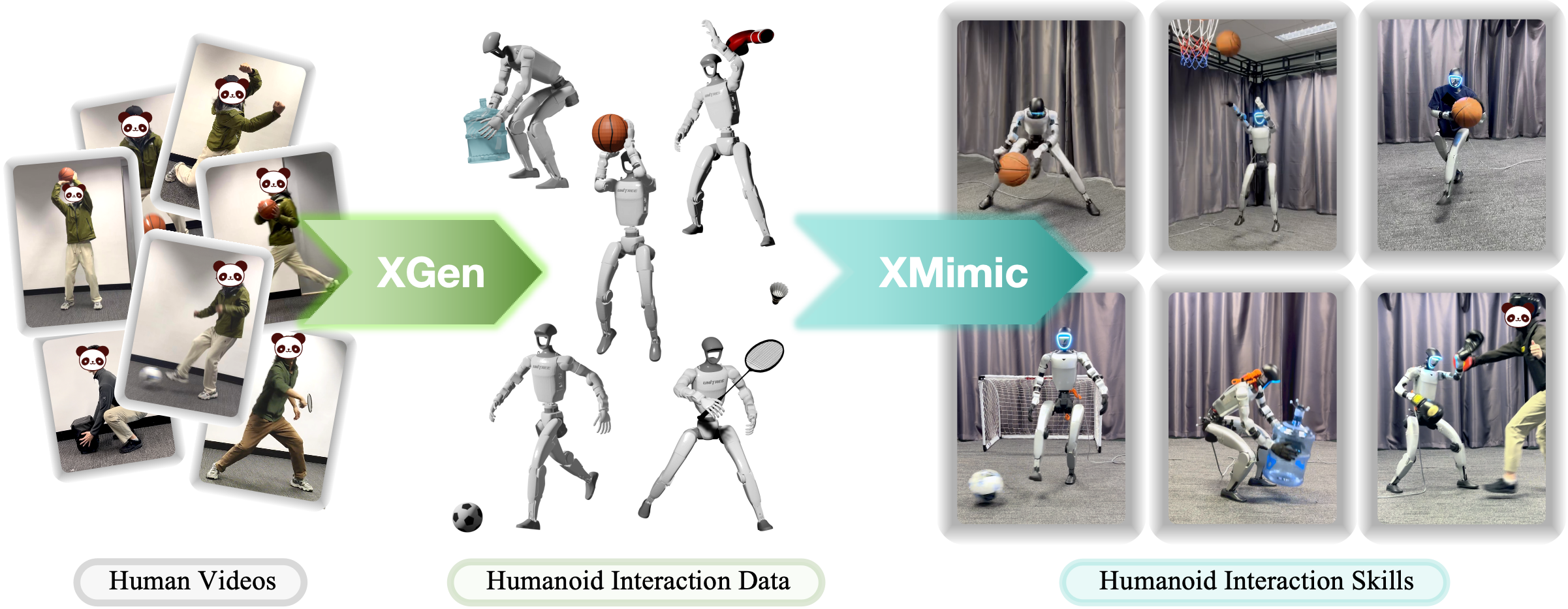}
    \caption{Breaking away from task‑specific engineering, we introduce a unified framework that compiles human video into generalizable, real-world humanoid interaction skills, such as basketball dribbling, jumpshots, layups, consecutive football kicks, generalizable cargo pickup, and real-time counterattacks against a human, showing generalization far beyond the original human video.}
    \label{fig: teaser}
\end{center}


\begin{abstract}
    Enabling humanoid robots to perform agile and adaptive interactions has long been a core challenge in robotics. Current approaches are bottlenecked by either the scarcity of realistic interaction data or the need for meticulous, task-specific reward engineering, which limits their scalability. To narrow this gap, we present HumanX, a unified framework that compiles human video into generalizable, real-world interaction skills for humanoids, without per-task design. HumanX integrates two co‑designed components: XGen, an anchor-centric data generation pipeline that yields physically plausible robot interaction data from monocular video and augments it to create a diverse, scalable training dataset; and XMimic, an interaction-prioritized learning framework to achieve generalizable interaction skills from seconds of demonstration.
    Evaluated across five distinct domains—basketball, football, badminton, cargo pickup, and reactive fighting—HumanX successfully acquires 9 different skills and transfers them zero‑shot to a physical Unitree G1 humanoid. The learned capabilities include complex maneuvers such as pump‑fake turnaround fadeaway jumpshots without any external perception, as well as interactive tasks like sustained human‑robot passing sequences over 10 consecutive cycles—learned from a single video demonstration. Our experiments show that HumanX achieves higher generalization success than prior methods, demonstrating a scalable pathway within a class of anchorable interactions for learning versatile, real‑world robot interactive skills.
\end{abstract}

\keywords{Humanoids, Whole-Body Control, Learning from Human Video}


\vspace{-0.5cm}
\section{Introduction}
\vspace{-0.3cm}
Humanoid robots share a morphological affinity with humans, offering the potential to operate seamlessly in human environments and interact with everyday objects. This inherent compatibility points to a vast, yet largely untapped resource: the rich diversity of skills demonstrated in human motion. However, unlocking this potential for robot learning remains a challenge. While behavior cloning (BC) offers a simple and unified training paradigm, it relies on large-scale, costly teleoperated demonstrations \cite{fu2024humanplus,bjorck2025gr00t,cheng2025open}. 
Although reinforcement learning (RL) combined with physics simulation can substantially reduce the demand for large quantities of high-quality demonstrations, it usually requires meticulously engineered, task-specific reward functions, limiting its scalability across diverse tasks \cite{Haarnoja2023LearningAS,Ma2025LearningCB,He2025LearningGP,wang2025physhsi,tennis,Xu2025LearningTB}.
Together, these bottlenecks have constrained the development of a general, scalable pipeline for acquiring humanoid interaction skills from human.

\vspace{-0.1cm}
To address these limitations, we introduce HumanX, a full‑stack framework that compiles human video into generalizable, real‑world interaction skills for humanoids, without per-task design. HumanX integrates two synergistic components: XGen, an \textcolor{black}{anchor-centered} data‑generation pipeline that yields diverse and physically plausible humanoid interaction data from monocular video while enabling scalable augmentation; and XMimic, a unified \textcolor{black}{interaction‑prioritized} learning framework that enables the mastery of generalizable interaction skills with only seconds of HOI data.

\vspace{-0.1cm}
While estimating human and object motion separately from monocular video is well-studied \cite{Loper2023SMPLAS,shen2024gvhmr,chen2025sam,Wen2023FoundationPoseU6,Sun2022OnePoseOO}, naively combining such independent estimates often yields physically implausible results due to issues like occlusion and depth ambiguity \cite{Hasson2019LearningJR,wang2023physhoi}. XGen addresses this by fundamentally shifting the paradigm: it \textit{synthesizes} interaction trajectories governed by physical rules, rather than pursuing exact \textit{reconstruction}. This shift enables highly efficient data augmentation, allowing XGen to generate a broad distribution of physically consistent interaction trajectories from just a single video demonstration. Concretely, XGen operates in three stages: (1) extracting human motion and retargeting it to the robot; (2) \textcolor{black}{interaction synthesis: anchor-object configuration, anchor-based object trajectory synthesis, and contact refinement}; (3) data augmentation through object geometry scaling and trajectory variation to maximize coverage for improved generalization.

\vspace{-0.1cm}
\textcolor{black}{
Instead of pursuing task-specific solutions \cite{he2025viral,kalaria2025dreamcontrol,su2025hitter,xue2025opening}, XMimic adopts an unified imitation paradigm for learning interaction skills. However, separately imitating body and object motions often results in weak interaction robustness and generalization \cite{zhao2025resmimic, yang2025omniretarget}. We identify that the core of interaction skills lies in the relative motion patterns between robot and object, which we formulated as \textit{Interaction Vectors (IVs)}. Thus, XMimic introduce three key designs to prioritize interaction learning: (1) \textit{IV Imitation Rewards} that incentivize relative body–object motion, enabling robust and generalizable interactions; (2) \textit{IV-based Termination}, which executes early termination \cite{DeepMimic} upon interaction failure, compelling the policy to prioritize interaction success under disturbances; and (3) \textit{Interaction Perturbation}, include perturbing the initialization of the robot-object state and applying random external forces to the robot and object during training, forcing the policy to adapt and generalize.
Together, these components enables robust and generalizable interaction skill from a single reference demonstration. This learning capability is further strengthened by XGen's data augmentation, achieves generalization far beyond the original video.}

\vspace{-0.1cm}
We evaluate HumanX on 9 diverse loco‑manipulation and interaction skills—spanning basketball, football, badminton, cargo handling, and robot-human fighting—on a Unitree G1 humanoid. The system demonstrates two practical deployment ways: (1) Without any explicit external sensing, it executes basketball skills including dribbling, layups, and complex pump‑fake turnaround fadeaways, with an average success rate over 80\%. (2) With object sensing from a MoCap system, it achieves sustained closed‑loop interactions, including over 10 consecutive human‑robot basketball passes and football kicks, along with reliable pickup of randomly placed objects. Notably, each skill is learned from a single video demonstration, highlighting the strong generalization capability of our approach. Beyond mimicry, the policies exhibit adaptive behaviors: if a human removes a carried object and sets it down, the robot autonomously walks to and regrasps it
—demonstrating real‑time reaction rather than fixed motion replay. Quantitatively, HumanX achieves significantly higher generalization success than existing methods, establishing a scalable pathway within a class of anchorable interactions for acquiring versatile interactive skills from human videos.

\section{Related Work}

\vspace{-0.3cm}
\paragraph{Data Acquisition for Humanoid Loco-Manipulation}

Retargeting human motion to humanoids for robot learning has shown significant promise for agile, dynamic skills \cite{he2025asap,he2024omnih2o,pan2025agility,li2025clone,liao2025beyondmimic,zhang2025track,cheng2024expressive,ji2024exbody2,xie2026kungfubot,su2025hitter,zeng2025behavior,yin2025unitracker}. 
For instance, SFV \cite{peng2018sfv} estimates human pose from monocular video and enables simulated humanoids to perform complex acrobatics. 
SkillMimic \cite{wang2025skillmimic,wang2023physhoi} estimates both human and object motion from video to train diverse basketball skills in simulation. VideoMimic \cite{allshire2025visual} estimates human-scene interaction data from video and enables real-world humanoid-scene interaction through imitation. Meanwhile, GMR \cite{joao2025gmr} provides a general motion retargeting framework that maps human motion to various robot morphologies. Several recent methods explores retargeting human-object or human-scene interaction data to train loco-manipulation policies \cite{yang2025omniretarget,weng2025hdmi,wang2025physhsi}.
However, these methods either rely on high-quality human-object interaction (HOI) data for retargeting \cite{yang2025omniretarget}, or they are struggle to reconstruct physically plausible interactions from monocular video due to severe occlusion and depth ambiguity \cite{weng2025hdmi}.
Furthermore, these methods suffer from limited scalability, making it difficult to collect robot interaction data that adequately covers the distribution required for learning generalizable interaction skills.
In contrast, our framework overcomes these limitations by shifting the interaction data acquisition paradigm from reconstruction to synthesis. This paradigm shift not only acquires physically plausible interaction data but also supports data augmentation, enabling strong generalization when combined with our training framework.

\vspace{-0.3cm}
\paragraph{Reinforcement Learning for Humanoid Robots}

Reinforcement learning (RL) in physics simulation has emerged as a key paradigm for whole‑body humanoid control \cite{he2025asap,he2024omnih2o,li2025hold,liao2025beyondmimic,li2025amo,He2025LearningGP,Shao2025LangWBCLH,Xue2025AUA,Wang2025BeamDojoLA,Ben2025HOMIEHL,cheng2024expressive,li2025bfm,radosavovic2024humanoid}. Early RL approaches for humanoids largely focused on gait learning, typically requiring carefully designed task‑specific reward functions \cite{tedrake2004stochastic,heess2017emergence,li2021reinforcement,li2025reinforcement}. This reward‑engineering paradigm has also proven effective for a variety of other tasks, such as getting up \cite{He2025LearningGP,Huang2025LearningHS}, goalkeeping \cite{ren2025humanoid}, box carrying \cite{dao2024sim,wang2025physhsi}, and pick-and-place \cite{he2025viral,kalaria2025dreamcontrol}.
Inspired by the success of imitation learning in character animation \cite{DeepMimic,amp,luo2023perpetual,xu2023composite}, retargeting human motion to humanoids and applying imitation rewards has enabled robots to acquire diverse locomotion skills—such as parkour \cite{zhuang2024humanoid,zhang2025wococo}, martial arts \cite{xie2026kungfubot}, jumping \cite{he2025asap,huang2025towards}, and even generalizable whole‑body motion tracking \cite{yin2025unitracker,zeng2025behavior,chen2025gmt,ze2025twist,li2025clone,pan2025agility}—through unified imitation rewards. Extending this imitation‑based paradigm to interaction has seen initial progress in simulation \cite{wang2023physhoi,wang2025skillmimic,xu2025intermimic,tessler2025maskedmanipulator}. 
\textcolor{black}{Recent research has explored imitation learning to equip real humanoid robots with interaction skills \cite{weng2025hdmi,yang2025omniretarget}. However, these methods typically imitate isolated object and body motions, causing the policy to easily overfit to isolated body motion patterns, which reduces interaction generalization and robustness. 
In contrast, our reward function and training mechanism are designed based on the interaction-first principle, achieving interaction performance far exceeding that of concurrent works.}

\vspace{-0.2cm}
\section{XGen}
\begin{figure*}[t]
  \centering  
  \vspace{-0.5cm}
  \includegraphics[width=\linewidth]{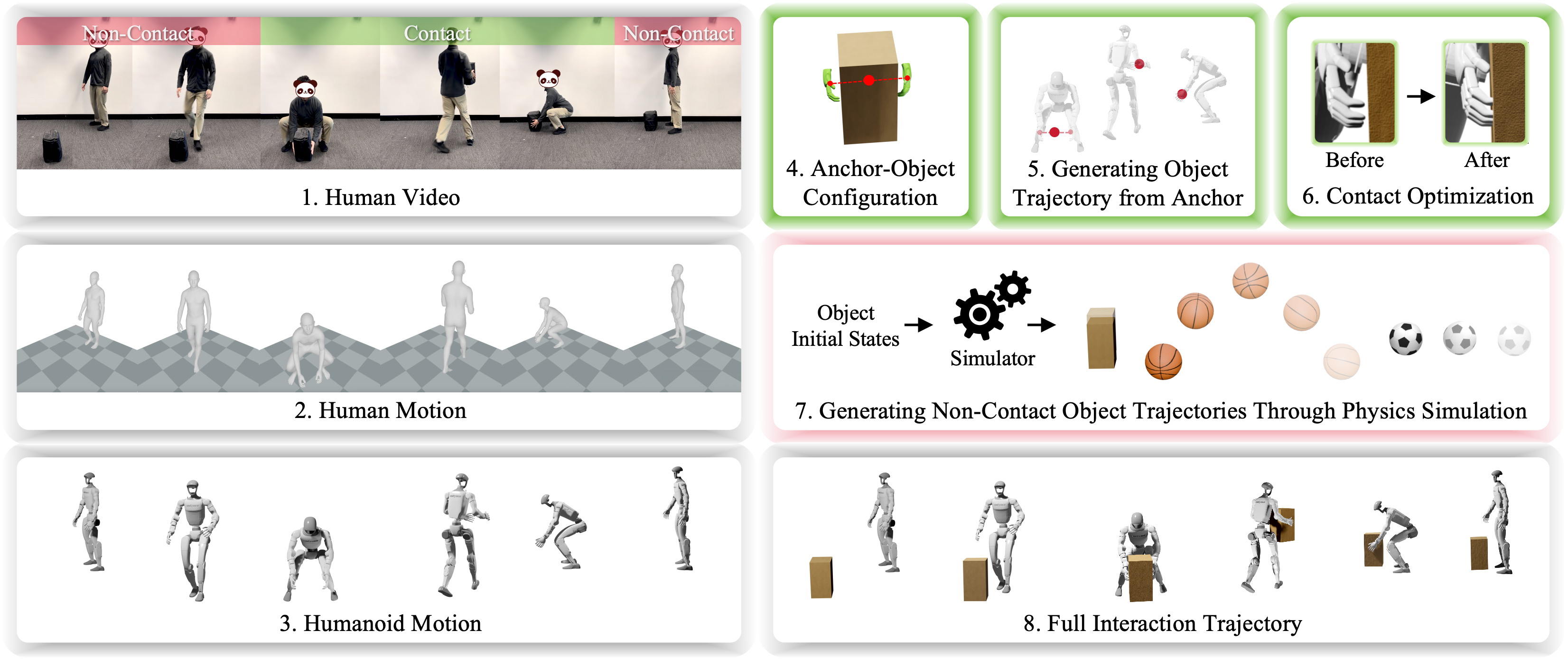}
  \caption{Overview of \textbf{XGen}. The pipeline begins by estimating SMPL‑based human motion from video and retargeting it to the humanoid’s morphology. The video is segmented into \textbf{\textcolor[RGB]{117,189,66}{contact}} and \textbf{\textcolor[RGB]{229,76,94}{non‑contact}} phases. For the contact phase, a body anchor (e.g., the midpoint between the two palms) is predefined.
  The object trajectory is then synthesized by transforming the object according to the anchor, followed by contact optimization to refine the robot poses. During the non‑contact phases, diverse object trajectories are generated via physical simulation. Complete interaction trajectories are obtained by concatenating and smoothly interpolating the phases. 
  } %
\label{fig: xgen}
\end{figure*}

\vspace{-0.3cm}
XGen is an \textcolor{black}{anchor-centric} data synthesis pipeline that generates physically plausible anchorable interactions from human demonstration videos. As illustrated in Fig.~\ref{fig: xgen}, it converts monocular human video into humanoid motion and synthesizes interaction based on anchor-object configuration and physical simulation. The pipeline further supports augmentation of object mesh, size, and trajectory to produce large-scale, diversified interaction data.

\vspace{-0.3cm}
\paragraph{Extracting Humanoid Motion from Human Video}

Given a monocular RGB video with $K$ frames, we first obtains an initial estimation of the 3D human pose sequence using off-the-shelf tools \cite{shen2024gvhmr}. The estimated 3D human pose in the $i$-th frame is defined as $\mathbf{h}_i = ( \mathbf{h}_i^{\text{root}},\; \mathbf{h}_i^{\text{joint}} )$, where $i = 1, \dots, K$.
$\mathbf{h}_i^{\text{root}} \in \mathbb{R}^6$ represents the 6D pose (3D position and 3D orientation) of the human root, and $\mathbf{h}_i^{\text{joint}} \in \mathbb{R}^{J \times 3}$ denotes the 3D rotations of $J$ SMPL \cite{Loper2023SMPLAS} joints.
\vspace{-0.1cm}
Subsequently, we use GMR \cite{joao2025gmr} to retarget the human pose sequence into a pose sequence of the target humanoid robot, which involves three core steps: keypoint alignment, skeleton scaling, and IK-based optimization. After retargeting, the robot pose sequence is denoted as $\mathbf{r}_i = (\mathbf{r}_i^{\text{root}},\; \mathbf{r}_i^{\text{joint}})$, where $i = 1, \dots, K$.
$\mathbf{r}_i^{\text{root}} \in \mathbb{R}^6$ is the 6D root pose, and $\mathbf{r}_i^{\text{joint}} \in \mathbb{R}^{N \times 1}$ represents the 1D rotations of $N$ joints.

\begin{figure*}[t]
  \centering
  \vspace{-0.45cm}
  \begin{minipage}[t]{0.55\textwidth}
      \centering  
      \includegraphics[width=\linewidth]{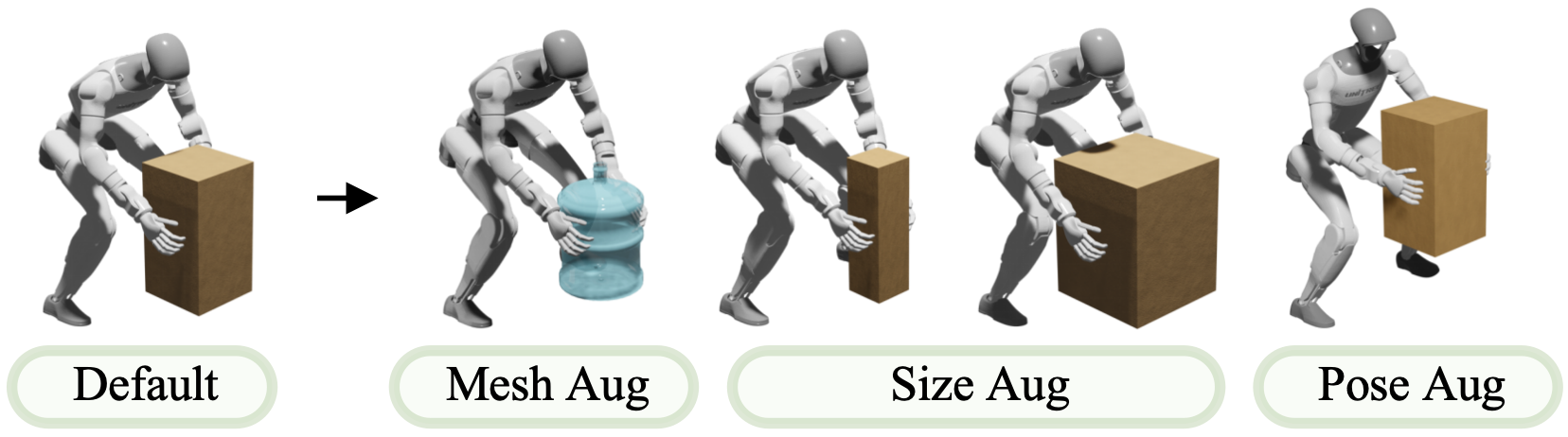}
      \caption{Data Augmentation for The Contact Phase.} %
      \label{fig: data augmentation c}
  \end{minipage}\hfill
  \begin{minipage}[t]{0.43\textwidth}
      \centering  
      \includegraphics[width=0.9\linewidth]{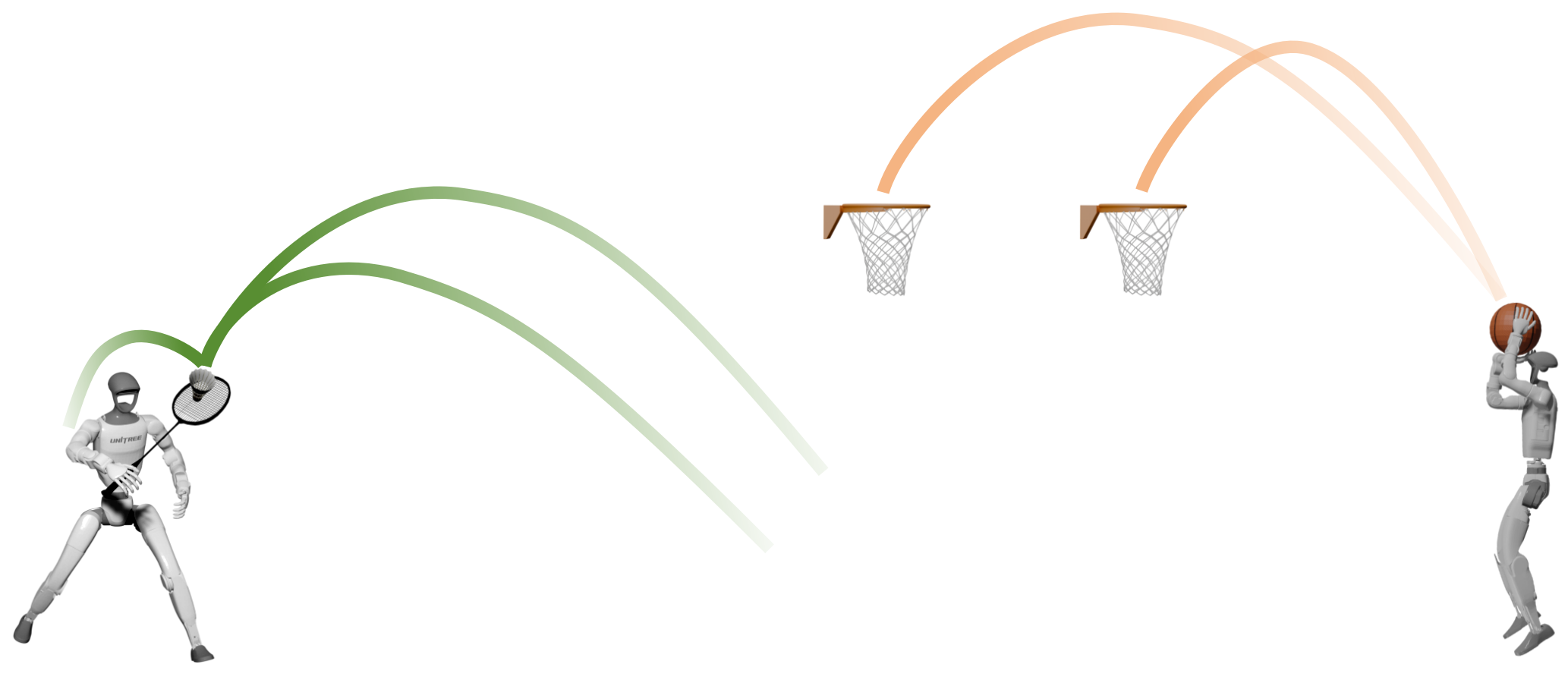}
      \caption{Aug for Non-Contact Phases.} %
    \label{fig: data augmentation nc}
  \end{minipage}
  \vspace{-0.3cm}
\end{figure*}

\vspace{-0.3cm}
\paragraph{Synthesizing Interaction via Anchor-Object Configuration}
We segment the data into contact and non-contact phases. As illustrated in Fig.~\ref{fig: xgen}, taking the example of box carrying, we annotate the video frames into: a non‑contact phase before contact begins ($t<t_s$), a contact phase ($t_s\leq t \leq  t_e$), and a final non‑contact phase after contact ends ($t>t_e$).

\vspace{-0.1cm}
For the contact phase, we consider the relative motion between a predefined body anchor and the object as the core of an interaction. This representation exhibits favorable cross‑embodiment properties, meaning the same anchor‑object relationship can be transferred across different morphologies (e.g., from human to humanoid) while preserving interaction semantics.
We primarily discuss two types of anchor definitions: (1) Using the midpoint between the two palms as the anchor, suitable for contact phases where the object is stably held with both hands, such as box carrying, shooting, and layups. (2) Using a specific body part as the anchor, suitable for contact phases involving a single-point interaction, such as hitting a shuttlecock or kicking a football ball.

\vspace{-0.1cm}
Once the anchor is defined, we can estimate the object’s mesh and its rotation \(\phi\) relative to the anchor at time \(t_s\) from the video frame \(\mathbf{v}_{t_s}\) using SAM-3D \cite{chen2025sam}. 
Alternatively, the mesh and the initial anchor‑object configuration can be manually defined, which also allows synthesizing interactions from videos where the object is not visibly present.
The anchor's trajectory is then derived from the robot motion sequence $\{\mathbf{r}_i\}$, and the corresponding object trajectory is obtained by preserving the relative transformation \(\phi\) throughout the anchor's motion.
To improve physical plausibility, the robot motion can be optimized under contact and force-closure objectives \cite{liu2021synthesizing, wang2022dexgraspnet,Wang2025LearningGH}, yielding refined robot poses \(\hat{\mathbf{r}}_t\) and corresponding object poses \(\mathbf{p}_t\) for each frame in the contact phase.

\vspace{-0.1cm}
In the non‑contact phase, object trajectories are synthesized using a physics simulator \cite{makoviychuk2021isaac}. Specifically: (1) After contact ends ($t>t_e$), the object is initialized in simulation with pose $\mathbf{p}_{t_e}$ and a predefined initial velocity, and its trajectory is recorded under simulation. This applies to actions such as basketball shooting, football kicking, or object placement. (2) Before contact begins ($t<t_s$), —e.g., when catching a ball—we reverse the process: starting from $\mathbf{p}_{t_s}$, we simulate the object backward in time and then reverse the sequence to obtain the pre‑contact trajectory. This allows accurate synthesis of motions such as a parabolic ball path into the hands. To ensure physical plausibility in the reversed simulation, object damping coefficients are inverted.

\begin{figure*}[t]
  \centering  
  \vspace{-0.5cm}
  \includegraphics[width=\linewidth]{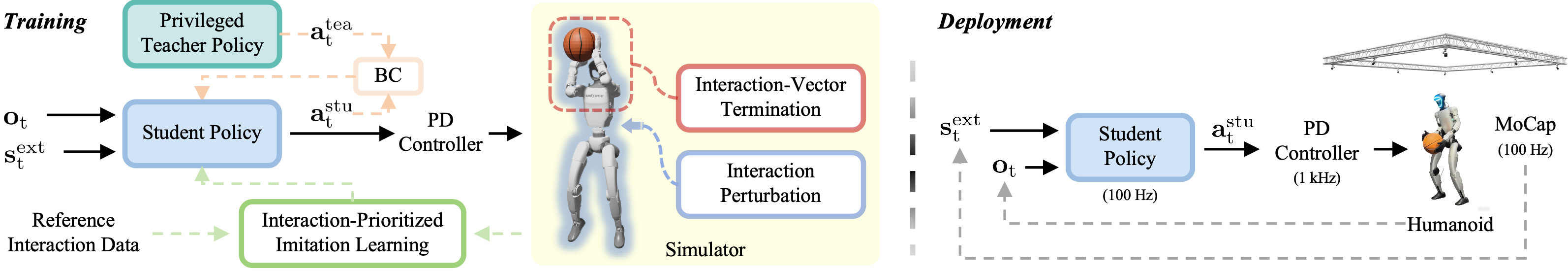}
  \caption{
  \textcolor{black}{\textbf{XMimic} is a unified framework for learning diverse interaction skills. To reinforce interaction and prevent overfitting to isolated body motions, it employs rewards that prioritize interaction and introduces interaction perturbations and interaction-vector termination during simulation. The training follows a two-stage pipeline: a privileged teacher policy first learns the skill using full state information, which then guides the training of a deployable student policy.
  }
  } 
\label{fig: xmimic}
\vspace{-0.3cm}
\end{figure*}

\vspace{-0.3cm}
\paragraph{Interaction Augmentation}

XGen supports data augmentation along multiple dimensions to increase the interaction diversity and data coverage: 
%
(1) We apply scaling to the object mesh or replace it with a different geometry during the mesh acquisition stage. The subsequent XGen synthesis process ensures that interactions remain physically plausible with the scaled or substituted object. This allows generating data for performing similar actions on different objects from a single demonstration video, as shown in Fig.~\ref{fig: data augmentation c}.
%
(2) The object trajectory within the contact phase can be augmented by applying simple geometric transformations—such as translation and scaling. The subsequent XGen pipeline ensures the physical plausibility of the interaction after augmentation. For example, from a single video demonstration of lifting a box, XGen can generate training data for lifting the same box from different heights, as illustrated in Fig.~\ref{fig: data augmentation c}.
(3) We enrich the diversity of object trajectories in the non-contact phase by introducing parametric randomization to the object's initial velocity in the physics simulation. 
For instance, from one demonstration of hitting a shuttlecock, XGen can produce data for hitting with different parabolic trajectories. Similarly, a single basketball shooting video can yield training data for making shots from various 
distances, as shown in Fig.~\ref{fig: data augmentation nc}.

\vspace{-0.2cm}
\section{XMimic}
\vspace{-0.3cm}
XMimic is a interaction-prioritized imitation learning framework that enables humanoids to acquire diverse interaction skills from humanoid-object interaction data. Fig.~\ref{fig: xmimic} provides an overview.

\vspace{-0.3cm}
\paragraph{Teacher-Student Training Architecture}
Our training process follows a two-stage teacher-student paradigm that first master individual skills with privilege information and then consolidate them into a unified deployable policy. 
%
%
%
Given the observation $\boldsymbol{s}_{t}$ as input, the policy output is parameterized as a Gaussian distribution $\boldsymbol{\pi}(\boldsymbol{a}_t\mid\boldsymbol{s}_t) \sim \mathcal{N}(\boldsymbol{\phi}_{\boldsymbol{\pi}}(\boldsymbol{s}_{t}), \boldsymbol{\Sigma}_{\boldsymbol{\pi}})$,
where $\boldsymbol{\phi}_{\boldsymbol{\pi}}$ is a MLP that predicts mean of the action distribution. The covariance matrix $\boldsymbol{\Sigma}_{\boldsymbol{\pi}}$ is learnable. The resulting action $\boldsymbol{a}_{t} \in \mathbb{R}^n$ (where $n$ is the number of robot DoFs) is then transformed into joint torques via a PD controller.
Given $n$ skill patterns and their corresponding datasets \( \{\mathcal{D}_1, \dots, \mathcal{D}_n\} \) generated by XGen, we train a teacher policy \( \pi^i_\text{tea} \) on each dataset \( \mathcal{D}_i \). 
The training procedure for a single teacher is as follows. A trajectory clip is sampled from its dedicated dataset, and the humanoid along with objects are initialized according to the first frame of the clip.
At each timestep $t$, the policy receives a privileged state observation $\boldsymbol{s}_{t} = \{\boldsymbol{o}_{t},\boldsymbol{o}_{t}^{priv},\boldsymbol{s}_{t}^{ext}\}$, which comprises proprioception $\boldsymbol{o}_{t}$, privileged body information $\boldsymbol{o}_{t}^{priv}$, and object state $\boldsymbol{s}_{t}^{ext}$. The policy then samples an action $\boldsymbol{a}_{t}$, which is executed in the physics simulator. Subsequently, the reward \( r_{t} \) is computed. The network parameters $\boldsymbol{\phi}_{\boldsymbol{\pi}}$ of the teacher policy are optimized using PPO \cite{schulman2017proximal} to maximize the expected cumulative reward.

\vspace{-0.1cm}
The student policy is trained on the combined dataset \( \mathcal{D} = \bigcup_i \mathcal{D}_i \) following similar procedure to the teacher, with proprioception and optional object observations. The training objective is extended to combine the RL rewards with a behavior cloning loss that distills the pre‑trained teachers. An asymmetric Actor-Critic setting \cite{pinto2018asymmetric} is applied to further provide privileged guidance for the student.

\vspace{-0.3cm}
\paragraph{Interaction-Prioritized Imitation Learning}
\textcolor{black}{
The essence of imitating an interaction skill lies not in replicating isolated body movements, but in learning relative body-object interaction patterns that are robust to perturbations. We capture this relational structure via Interaction Vectors (IVs). By imitating IVs from reference data and using IV tracking error to trigger early termination \cite{DeepMimic}, our approach significantly enhances the robustness and generalization of learned interactions. Concretely, we augment the anchor body with additional tracking points—such as the fingertips or the edge points of a racket—which, together with standard body keypoints, form an extended set of body points. The vectors from these points to the object’s center are defined as the Interaction Vectors (IVs). Interaction quality is measured by the consistency between the IVs in reference data and those in simulation. Thanks to their relative nature, IVs provide dense, fine‑grained guidance even under perturbation, whereas isolated object or body tracking rewards may fail in such cases. The complete reward function combines the IV imitation reward with isolated body/object tracking rewards, a contact reward, and regularization terms. We further employ IV error to detect interaction failure and trigger early termination, which we term IV-based Termination (IVT). IVT helps prevent the policy from overfitting to isolated body imitation. We apply IVT probabilistically to provide a tradeoff from strict to lenient. To further improve robustness and generalization, we introduce a set of Interaction Perturbations (IP) during training, including: (1) perturbations to the robot’s root pose, joint angles, and object pose at episode initialization, and (2) applying random external forces to the object (only during contact phases) and to the robot. Through the combined use of IV reward and IVT, the policy learns to compensate for these perturbations, achieving strong interaction performance that generalizes beyond the original demonstration. 
}

\begin{table*}[t]
    \vspace{-0.4cm}
    \caption{\textbf{Main Simulation Results}. SR, $E_o$, and $E_h$ 
    measure the success rate on the original data, the object position tracking error, and the key-body position tracking error, respectively, while GSR measures the success rate of skill generalization.
    }
\centering
\resizebox{\linewidth}{!}{%
\begin{tabular}{l|cccc|cccc|cccc}
\hline
\addlinespace[0.15em]

\multirow{2}{*}{Method} & \multicolumn{4}{c}{\textit{Basketball Catch-Shot}} & \multicolumn{4}{c}{\textit{Badminton Hitting}} & \multicolumn{4}{c}{\textit{Cargo Pickup}}\\

\addlinespace[0.15em]
\cline{2-13}
\addlinespace[0.15em]

 & SR$^\uparrow$ & GSR$^\uparrow$ & ${E_{o}}^\downarrow$ (m) & ${E_{h}}^\downarrow$ (cm)& SR$^\uparrow$ & GSR$^\uparrow$ & ${E_{o}}^\downarrow$ (m)& ${E_{h}}^\downarrow$ (m)& SR$^\uparrow$ & GSR$^\uparrow$ & ${E_{o}}^\downarrow$ (cm)& ${E_{h}}^\downarrow$ (cm)
\\

\addlinespace[0.15em]
\hline
\addlinespace[0.15em]

SkillMimic \cite{wang2025skillmimic} & 
0.0\% & 0.0\% & 1.25 \textcolor{gray}{\text{\tiny($\pm$ 0.26)}}& 10.3 \textcolor{gray}{\text{\tiny($\pm$ 1.37)}}& 
68.2\% & 30.9\% & 2.94 \textcolor{gray}{\text{\tiny($\pm$ 0.91)}} & 0.48 \textcolor{gray}{\text{\tiny($\pm$ 0.13)}}& 
0.4\% & 0.0\% & 25.1 \textcolor{gray}{\text{\tiny($\pm$ 0.06)}} & 28.3 \textcolor{gray}{\text{\tiny($\pm$ 2.13)}} \\

OmniRetarget \cite{yang2025omniretarget} & 
0.0\% & 0.0\% & 1.18 \textcolor{gray}{\text{\tiny($\pm$ 0.15)}} & 7.83 \textcolor{gray}{\text{\tiny($\pm$ 0.95)}}& 
75.8\% & 20.4\% & 1.25 \textcolor{gray}{\text{\tiny($\pm$ 1.15)}}& 0.15 \textcolor{gray}{\text{\tiny($\pm$ 0.10)}} & 
0.0\% & 0.0\% & 10.8 \textcolor{gray}{\text{\tiny($\pm$ 0.00)}} & 13.3 \textcolor{gray}{\text{\tiny($\pm$ 1.57)}} \\

HDMI \cite{weng2025hdmi} & 
53.1\% & 2.4\% & 0.70 \textcolor{gray}{\text{\tiny($\pm$ 0.11)}} & 10.8 \textcolor{gray}{\text{\tiny($\pm$ 2.41)}}&
90.0\% & 25.3\% & 1.64 \textcolor{gray}{\text{\tiny($\pm$ 0.66)}} & 0.07 \textcolor{gray} {\text{\tiny($\pm$ 0.02)}} & 
95.8\% & 1.8\% & 1.79 \textcolor{gray}{\text{\tiny($\pm$ 0.22)}} & 13.1 \textcolor{gray}{\text{\tiny($\pm$ 1.28)}} \\

\addlinespace[0.15em]
\hline
\addlinespace[0.15em]

XMimic (Base)& 
93.4\% & 4.9\% & 0.69 \textcolor{gray}{\text{\tiny($\pm$ 0.14)}} & 6.36 \textcolor{gray}{\text{\tiny($\pm$ 1.51)}}& 
97.1\% & 41.6\% & 1.32 \textcolor{gray}{\text{\tiny($\pm$ 0.44)}}& 0.12 \textcolor{gray}{\text{\tiny($\pm$ 0.07)}} & 
100\% & 50.9\% & 1.56 \textcolor{gray}{\text{\tiny($\pm$ 0.30)}} & 6.58 \textcolor{gray}{\text{\tiny($\pm$ 0.76)}} \\

+ IP & 
86.8\% & 13.5\% & 0.70 \textcolor{gray}{\text{\tiny($\pm$ 0.08)}} & 7.46 \textcolor{gray}{\text{\tiny($\pm$ 0.46)}}& 
94.0\% & 60.5\% & 1.03 \textcolor{gray}{\text{\tiny($\pm$ 0.57)}}& 0.10 \textcolor{gray}{\text{\tiny($\pm$ 0.05)}} & 
100\% & 94.5\% & 1.62 \textcolor{gray}{\text{\tiny($\pm$ 0.03)}} & 10.5 \textcolor{gray}{\text{\tiny($\pm$ 1.23)}} \\

+ IVT & 
93.4\% & 10.9\% & 0.83 \textcolor{gray}{\text{\tiny($\pm$ 0.09)}} & 6.97 \textcolor{gray}{\text{\tiny($\pm$ 0.38)}}&
100\% & 67.2\% & 1.33 \textcolor{gray}{\text{\tiny($\pm$ 0.67)}}& 0.10 \textcolor{gray}{\text{\tiny($\pm$ 0.06)}} &
99.6\% & 94.3\% & 1.51 \textcolor{gray}{\text{\tiny($\pm$ 0.29)}} & 9.62 \textcolor{gray}{\text{\tiny($\pm$ 1.13)}} \\

+ Data Aug & 
77.6\% & 60.6\% & 0.76 \textcolor{gray}{\text{\tiny($\pm$ 0.11)}} & 8.34 \textcolor{gray}{\text{\tiny($\pm$ 0.91)}}&
100\% & 84.4\% & 0.98 \textcolor{gray}{\text{\tiny($\pm$ 0.65)}}& 0.10 \textcolor{gray}{\text{\tiny($\pm$ 0.05)}} &
/ & / & / & / \\

+ Tea-Stu & 
86.8\% & 64.7\% & 0.85 \textcolor{gray}{\text{\tiny($\pm$ 0.10)}} & 7.83 \textcolor{gray}{\text{\tiny($\pm$ 0.95)}}&
100\% & 90.6\% & 0.90 \textcolor{gray}{\text{\tiny($\pm$ 0.46)}}& 0.09 \textcolor{gray}{\text{\tiny($\pm$ 0.06)}} &
99.3\% & 96.3\% & 1.38 \textcolor{gray}{\text{\tiny($\pm$ 0.25)}} & 8.67 \textcolor{gray}{\text{\tiny($\pm$ 0.91)}} \\

\bottomrule
\end{tabular}
}
\label{tab: main}
\end{table*}

\begin{figure*}[t]
  \centering
  \vspace{-0.3cm}
  \begin{minipage}[t]{0.48\textwidth}
      \centering  
      \includegraphics[width=\linewidth]{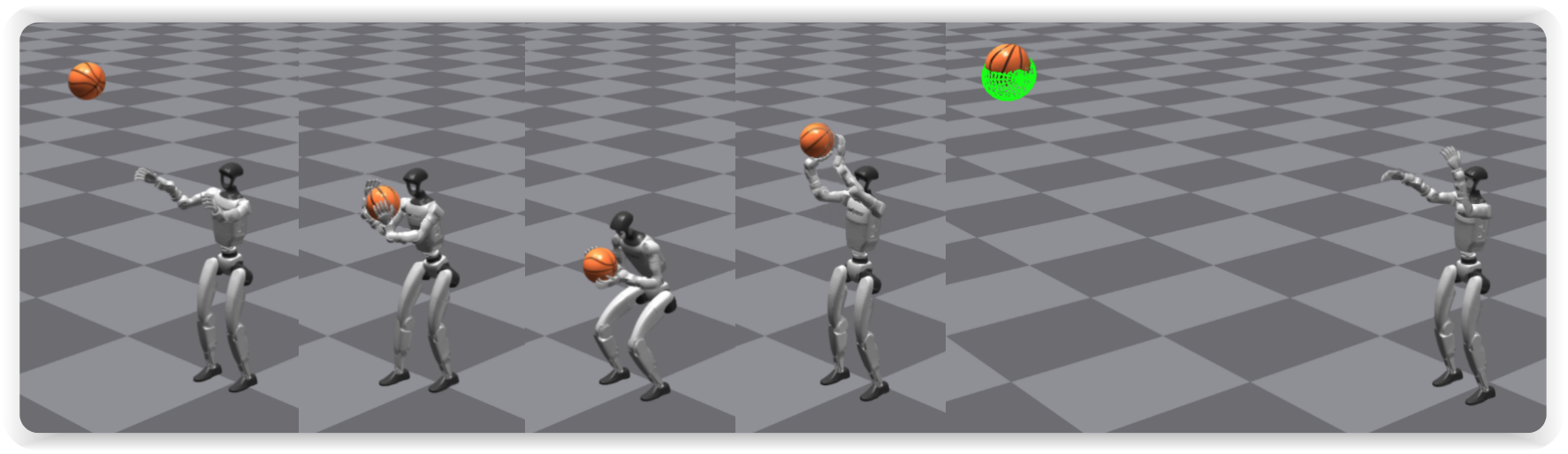}
      \caption{\textbf{Simulation Results on Basketball Catch-Shot}. XMimic generalizes to novel ball‑passing trajectories and target positions (green sphere) with natural interactions. 
      }
    \label{fig: sim result}
  \end{minipage}\hfill
  \begin{minipage}[t]{0.48\textwidth}
    \centering  
    \includegraphics[width=\linewidth]{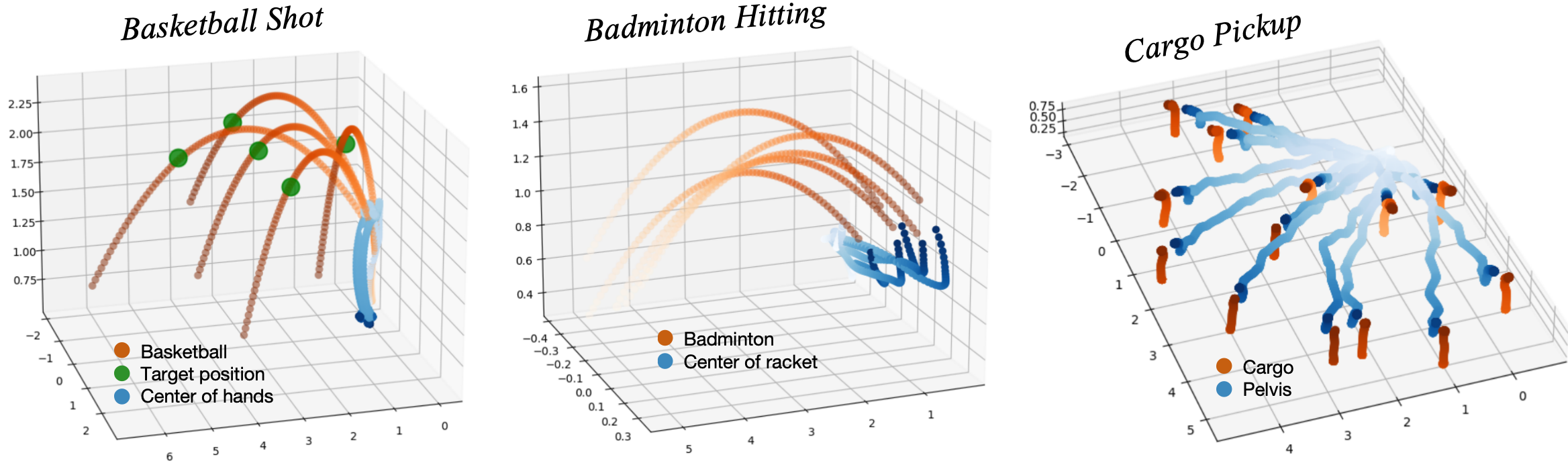}
    \caption{\textbf{Visualization of Generalization Performance in Simulation}. With HumanX, skills learned from only \textbf{one video} generalize to unseen object positions, trajectories, and goals.}
    \label{fig: generalization visulization}
  \end{minipage}
  \vspace{-0.3cm}
\end{figure*}

\vspace{-0.3cm}
\paragraph{Flexible Perception Schemes for Deployment}
XMimic supports two deployment schemes: (1) \textit{No External Perception (NEP) mode}, where object observations are removed during the student training, enabling the robot to rely solely on proprioception for dynamic interaction. This mode supports skills such as shooting, layups, dribbling, and complex maneuvers like pump‑fake turnaround jumpshots. 
(2) \textit{MoCap mode}, where the object observations are provided by a MoCap system. However, object tracking via MoCap often suffers from intermittent frame loss due to occlusion. To address this, we simulates frame loss of the object observations during the student training. This enables zero‑shot adaptation to real‑world MoCap streams with intermittent data loss.

\begin{figure*}[t!]
  \centering  
  \vspace{-0.4cm}
  \includegraphics[width=\linewidth]{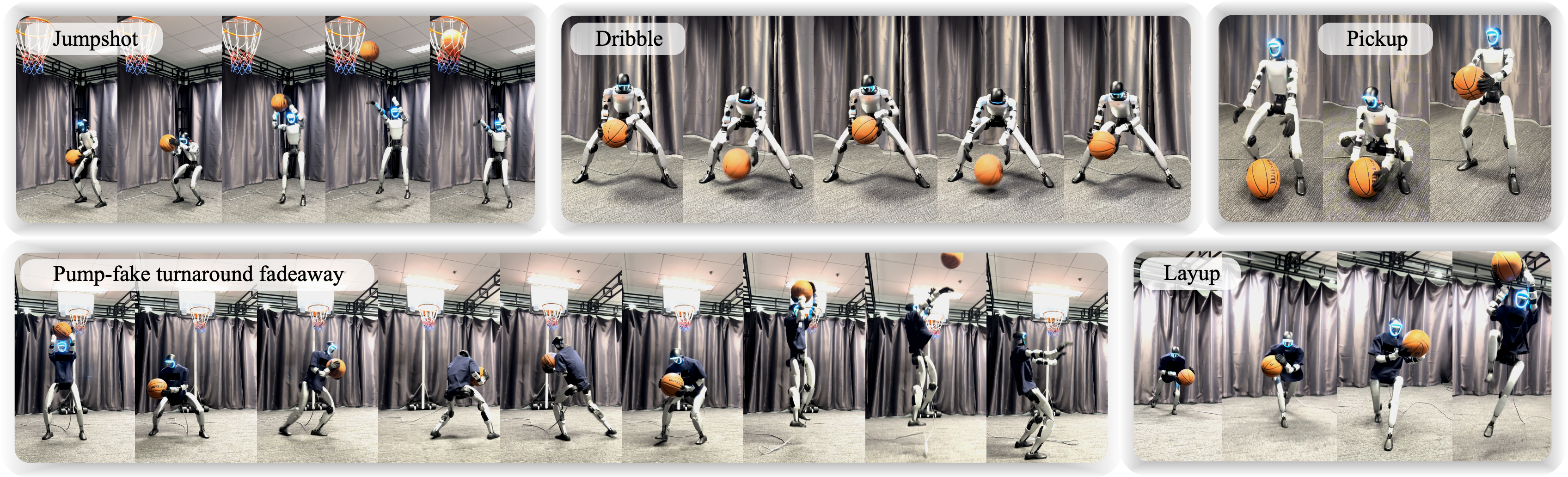}
  \caption{\textbf{Real Robot Experiment on Blind Basketball Skills}. 
  The policy relies solely on proprioception to control objects, achieving diverse, highly dynamic, and complex interactions.
  }
\label{fig: nep basketball results}
\end{figure*}

\begin{figure*}[t!]
  \centering  
  \vspace{-0.2cm}
  \includegraphics[width=\linewidth]{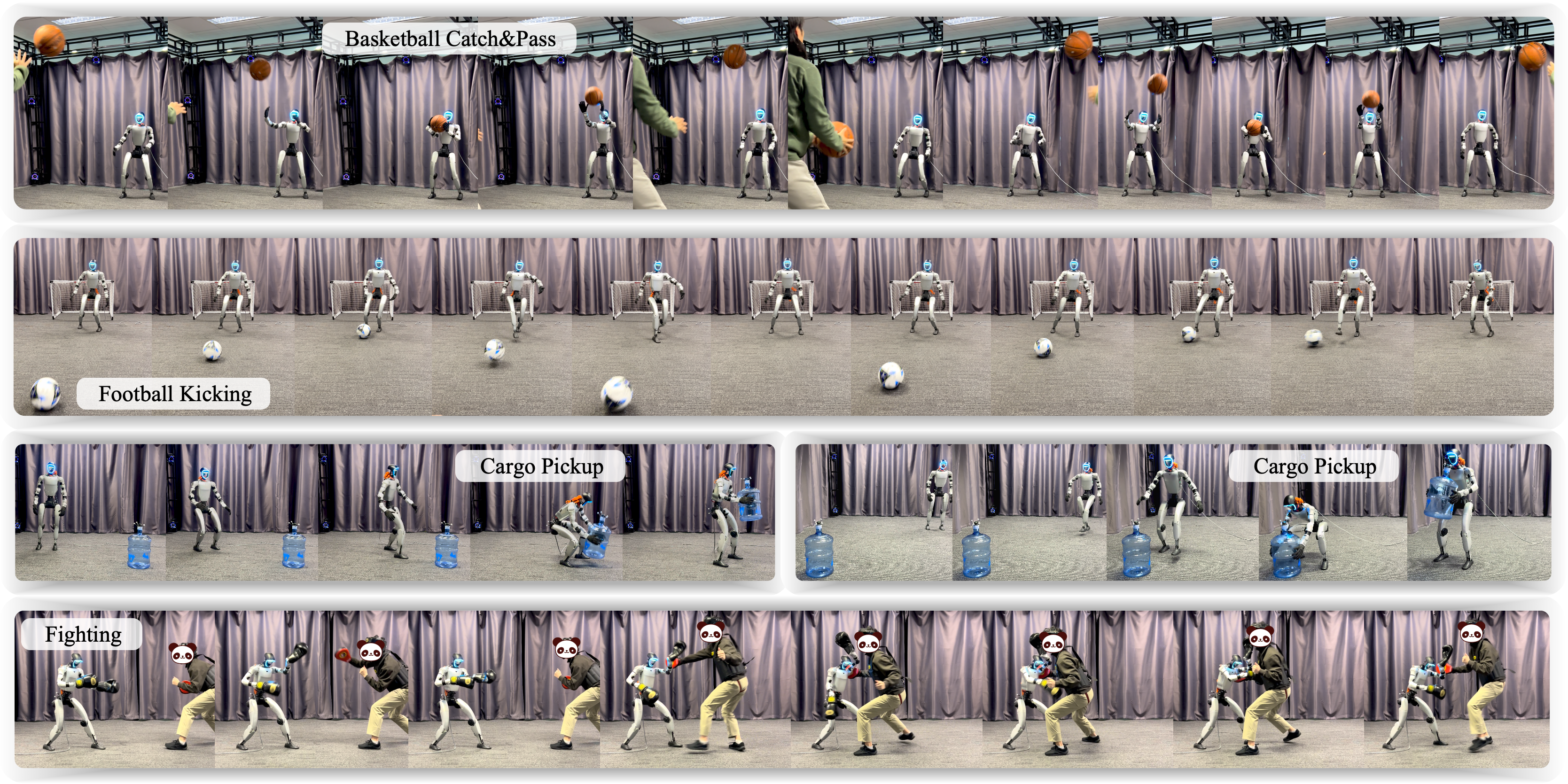}
  \caption{\textbf{Real Robot Experiment on MoCap-based Interaction Skills}. When utilizing MoCap system to perceive object or human motion, our method enables sustained interaction, demonstrating high precision, agility, robustness, and generalization capability. Notably, each task shown here is learned from a single demonstration video without any task-specific reward.
  }
  \vspace{-0.3cm}
\label{fig: mocap results}
\end{figure*}

\vspace{-0.2cm}
\section{Experiments}

%

\vspace{-0.3cm}
\paragraph{Experimental Settings}

Video clips for XGen was captured using an iPhone 16. All training and simulation were conducted on the Isaac Gym platform \cite{makoviychuk2021isaac} using a single NVIDIA RTX 4090 GPU with 16,384 parallel environments. Each policy was trained for 20,000 iterations
unless otherwise specified.
Deployment was performed on a Unitree G1 humanoid robot. For MoCap-based experiments, we employed a Noitom optical motion capture system within a 5×5×2.6m space, using 14 cameras. The policy and mocap systems ran at 100 Hz, while the PD controller operated at 1 kHz.

%
\vspace{-0.3cm}
\paragraph{Simulation Evaluation and Ablation Study}
We conduct simulation experiments on three representative interaction tasks: (1) \textit{Basketball Catch-Shot}: catching a passed basketball and shooting it into a target hoop. Success is defined as the shot landing within 20 cm of the target hoop center. (2) \textit{Badminton Hitting}: striking a flying shuttlecock, with success measured by the hitting rate. (3) \textit{Cargo Pickup}: walking to and lifting a randomly placed cargo. Success requires the lifted cargo to reach within 10 cm of the target height.
We perform a series of ablation studies on our method, starting from our baseline (XMimic Base), and incrementally add: interaction perturbation (+IP), interaction-vector termination (+IVT), XGen data augmentation (+Data Aug), and the teacher-student scheme (+Tea-Stu). For comparison, we also evaluate on prior and concurrent imitation methods including SkillMimic \cite{wang2025skillmimic}, OmniRetarget \cite{yang2025omniretarget}, and HDMI \cite{weng2025hdmi}.
For each task, a single video is processed by XGen to generate one training clip. In the “+Data Aug” and “+Tea-Stu” settings, this clip is augmented by XGen to produce 50 interaction clips for training. 

\vspace{-0.1cm}
We report four metrics: object-position error (${E_{o}}$), key-body-position error (${E_{h}}$), success rate (SR) on the original demonstration, and generalization SR beyond the original demonstration (GSR).
Quantitative results are summarized in Tab.~\ref{tab: main}. \textcolor{black}{The multiplicative reward of SkillMimic is difficult to balance with regularization terms, leading to suboptimal performance. OmniRetarget simply sums the object and body tracking rewards, which biases the policy towards replicating isolated body motion patterns and yields weak interaction. HDMI incorporates an additional contact reward, thereby improving its ability to fit a single demonstration, but it suffers from poor generalization due to overfitting.
In contrast, XMimic (Base) yields superior interaction performance over these methods, validating the effectiveness of our Interaction-Vector reward. Subsequent ablation studies highlight significant gains in GSR, with our final model exceeding 80\% average GSR—approximately 8× higher than HDMI, further demonstrating the efficacy of XMimic's overall interaction-centric design.} 
Fig.~\ref{fig: sim result} illustrates simulated executions of our generalized \textit{Basketball Catch-Shot} skill. Fig.~\ref{fig: generalization visulization} visualizes the generalization ranges for each skill. Notably, this strong generalization capability is learned from only a single demonstration video per skill.

\begin{table}[t]
    \centering
    \vspace{-0.4cm}
    \caption{\textbf{Quantitative Results on Real Robot Experiments}.}
    \renewcommand{\arraystretch}{1.5}
    \resizebox{1\linewidth}{!}{
    \begin{tabular}{lclclc}
    \toprule[1.0pt]
    
    \textit{Basketball Catch-Pass} (MoCap)
    &41 / 50 & \textit{Reactive Fighting} (MoCap)
    &37 / 50 & \textit{Jumpshot} (NEP) &8 / 10
    \\
    \textit{Cargo Pickup} (MoCap)
    &43 / 50 & \textit{Dribble} (NEP) &8 / 10 & \textit{Basketball Pickup} (NEP)
    &10 / 10
    \\
    \textit{Football Kicking} (MoCap)
    &42 / 50 & \textit{Pump-fake} (NEP) &9 / 10 & \textit{Layup} (NEP) &7 / 10
    \\
    \bottomrule[1.0pt]
    \end{tabular}
    }
    \vspace{-0.3cm}
    \label{table: real robot experiment}
\end{table}

\begin{figure*}[t]
  \centering
  \begin{minipage}[t]{0.47\textwidth}
      \centering  
      \includegraphics[width=\linewidth]{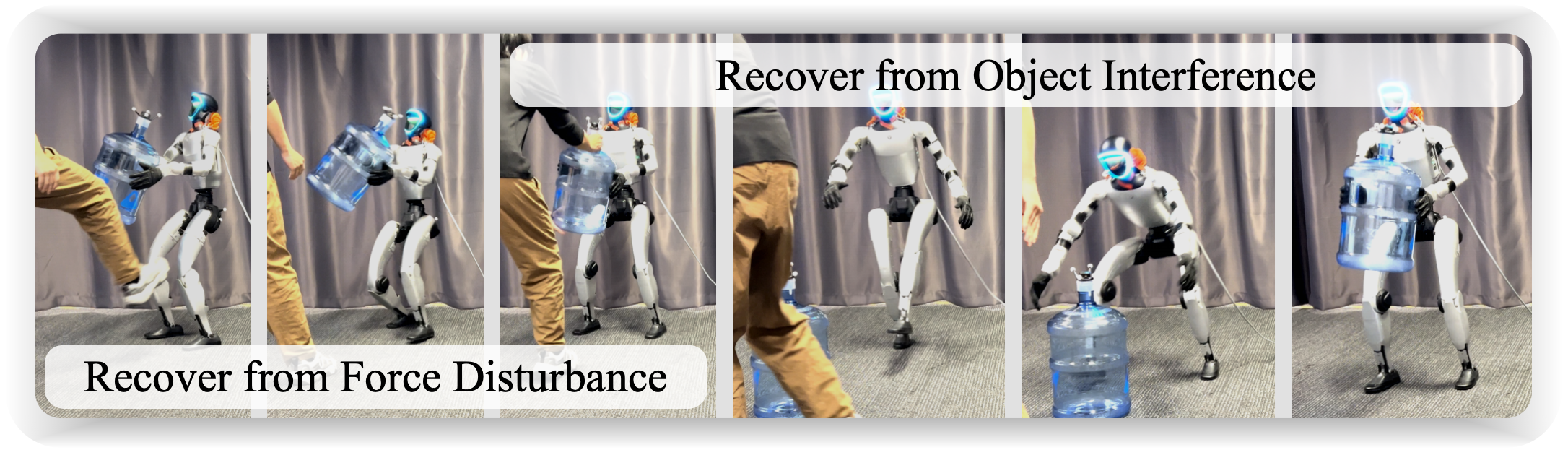}
      \caption{\textbf{Adaptive Behaviors}. During the execution of the Cargo Pickup skill, a researcher first kicks the robot forcefully, then takes the object from its hand and places it on the ground. The robot shows robust adaptation in response to such complex disturbances.
      }
    \label{fig: rubust}
  \end{minipage}\hfill
  \begin{minipage}[t]{0.51\textwidth}
      \centering  
      \includegraphics[width=\linewidth]{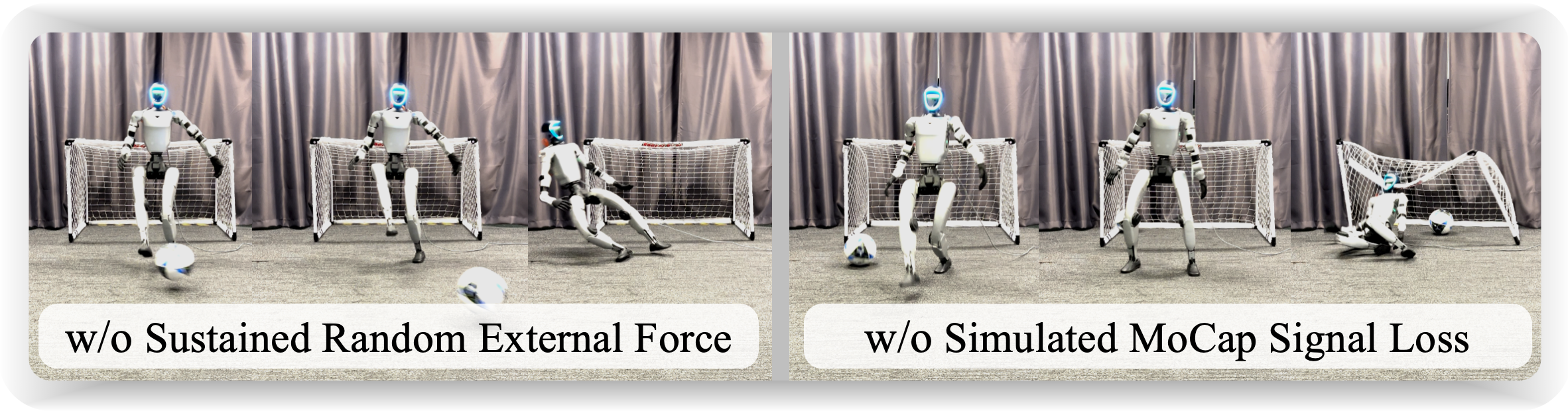}
      \caption{\textbf{Sim-to-Real Analysis}. (\textit{\textbf{Left}}) If the training does not include Interaction Perturbation, the robot may lose balance during highly dynamic interactions. (\textit{\textbf{Right}}) Without simulating MoCap signal loss during training, the robot may collapse when the object signal is lost during deployment.}
    \label{fig: sim2real analysis}
  \end{minipage}
  \vspace{-0.3cm}
\end{figure*}

\vspace{-0.3cm}
\paragraph{Real Robot Experiments}

We evaluate real‑world deployment under two perception schemes: \textit{NEP mode} and \textit{MoCap mode}. Each skill is trained based on a single human video.

\vspace{-0.1cm}
\textit{NEP Mode.}
We test five basketball skills in NEP mode: \textit{Jumpshot}, \textit{Dribble}, \textit{Pickup}, \textit{Layup}, and \textit{Pump‑fake turnaround fadeaway}. Each skill starts with the ball in hand (or on the floor for Pickup). A trial is successful if the robot completes the entire sequence without dropping the ball and remains balanced. Over 10 trials per skill (Tab.~\ref{table: real robot experiment}), the policy achieves high success rates, demonstrating reliable proprioceptive control for diverse interaction behaviors (Fig.~\ref{fig: nep basketball results}).

\vspace{-0.1cm}
\textit{MoCap Mode.}
We evaluate four interactive tasks using MoCap: \textit{Cargo Pickup}, sustained \textit{Basketball Catch‑and‑Pass} and \textit{Football Kicking} with a human partner, and \textit{Reactive Fighting} (blocking and countering human punches). Success rates over 50 trials per skill are reported in Tab.~\ref{table: real robot experiment}.
Our method enables prolonged, closed‑loop interaction. For instance, in basketball, the robot can execute over 10 consecutive successful catch‑and‑pass cycles with a human partner, maintaining stability even if the ball is dropped and seamlessly resuming when the ball is returned. Similarly, in football, it achieves over 14 consecutive successful return kicks despite variability in human passes. This consistent performance under human‑induced uncertainty strongly evidences the policy’s generalization capability.
The results visualized in Fig.~\ref{fig: mocap results} and Fig.~\ref{fig: rubust} demonstrate that the acquired skills extend far beyond simple mimicry. They show adaptive closed‑loop execution, robustness to perturbations, and the ability to generalize within interactive scenarios, validating the effectiveness of our HumanX system.
Finally, Fig.~\ref{fig: sim2real analysis} illustrates two key factors critical for deployment stability.

\vspace{-0.3cm}
\section{Conclusion, Limitation, and Future Extension} 
\label{sec:conclusion}
\vspace{-0.3cm}
\textcolor{black}{This paper presents HumanX, a full‑stack framework that compiles monocular human video into agile and generalizable humanoid interaction skills. With HumanX, we demonstrate a range of interactive capabilities that are difficult to achieve with prior methods. Nevertheless, the framework still has several limitations. First, while XGen is effective for synthesizing tasks involving stable grasps and point‑contact interactions—which constitute a large portion of practical manipulation—it does not yet support in‑hand reorientation motions. Second, due to limits in data reconstruction and simulation techniques, our method cannot currently handle interactions with deformable or cuttable objects, such as dumpling wrapping, apple peeling, or garment dressing. Third, our implementation does not employ dexterous hands, though integrating mature hand-object grasp synthesis techniques \cite{wang2023dexgraspnet,chen2025dexonomy} with XGen’s anchor‑centered pipeline is a promising direction. Beyond data synthesis, the current motion‑capture‑based perception system struggles to perceive diverse objects and is costly to deploy. In the future, we plan to investigate egocentric visual perception, which will help scaling a single policy to acquire a wide range of interaction skills with cross‑object generalization.}


\clearpage
\acknowledgments{We sincerely thank Weishuai Zeng, Tao Huang, Feiyu Jia, Huayi Wang, Xiao Chen, Junli Ren, and Wenjun Liu for their help, discussions, and support throughout the experiments. We also express our gratitude to Unitree Robotics for providing technical support on the robot, and to Noitom Robotics for their technical assistance with the motion capture equipment. This work was supported by the HKUST-BYD Joint Laboratory and Shanghai Artificial Intelligence Laboratory.}


\bibliography{example}  


\clearpage

\section{Appendix}

\subsection{Evaluation on Multi-Pattern Interaction Skills}

\begin{figure}[t]
  \centering  
  \includegraphics[width=\linewidth]{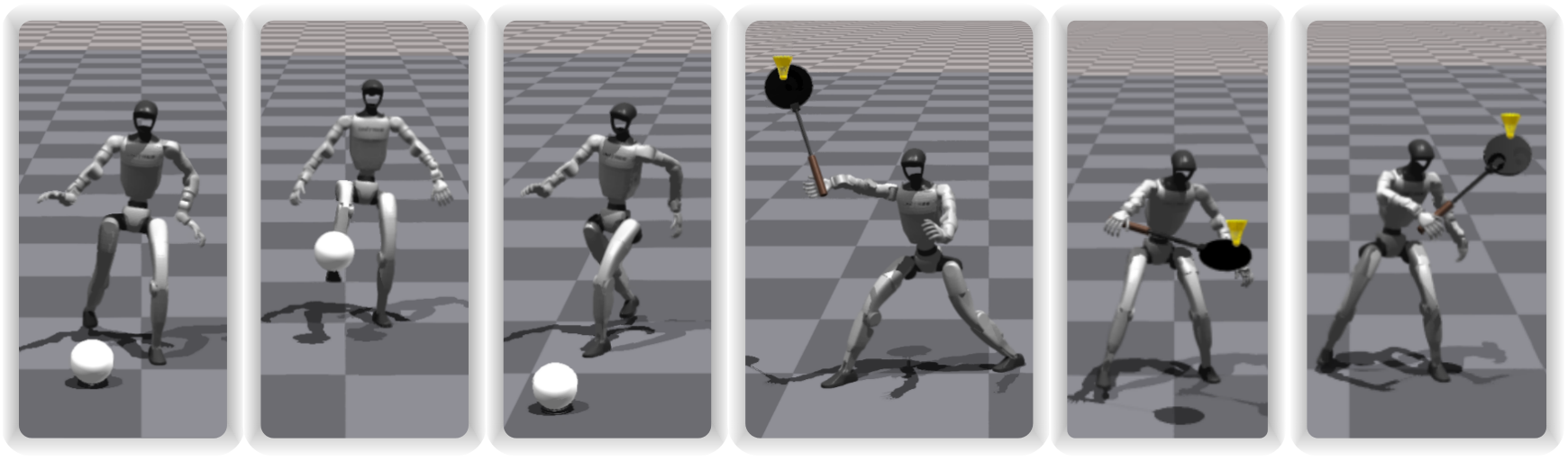}
  \caption{\textbf{Diverse Skill Patterns}. XMimic supports learning multiple interaction patterns for a single skill, allowing the policy to autonomously select the most suitable pattern in response to object state. (\textit{\textbf{Left}}): football‑kicking patterns. (\textit{\textbf{Right}}): badminton‑hitting patterns.
  }
  \vspace{-0.2cm}
\label{fig: multi-skill}
\end{figure}

To assess whether a single student policy can learn diverse skill patterns, we test on two representative tasks: \textit{Football Kicking} and \textit{Badminton Hitting}. For each task, three distinct human demonstration videos are processed by XGen and augmented for training. An ablation study confirms the critical role of the teacher-student scheme in this setting. Quantitative and qualitative results (Tab.~\ref{table: multi-skill} and Fig.~\ref{fig: multi-skill}) show that XMimic successfully learns multiple patterns, with the teacher-student framework yielding greater benefits in multi-pattern learning than in single-pattern scenarios.

\subsection{GSR Evaluation Setting}
For evaluating the Generalization Success Rate (GSR), the test cases are sampled from a distribution beyond the original data. For Basketball Catch-and-Shot, the ball's initial position is perturbed by ±0.3m (uniform distribution). This creates novel passing trajectories and requires shooting to a new target hoop location simultaneously.
For Badminton Hitting, the shuttlecock's initial position is perturbed by ±0.3m (uniform distribution). For Cargo Pickup, the object is randomly placed within a semicircular area of 3m radius in front of the robot's initial orientation.

\begin{table}[t]
    \centering
    \caption{\textbf{Evaluation on Multi-Pattern Interaction Skills in Simulation}. Each skill contains three distinct interaction patterns from three human video.
    }
    \vspace{0.1cm}
    \renewcommand{\arraystretch}{1.5}
    \resizebox{0.7\linewidth}{!}{
    \begin{tabular}{lcccc}
    \toprule[1.0pt]
    \textbf{Method} &   \quad\quad\textit{Football Kicking}\quad\quad\quad &  \quad\quad\textit{Badminton Hitting}\quad\quad\quad\\
    
    \midrule[0.6pt]
    w/o Tea-Stu
    &  74.2\% GSR  & 52.4\% GSR \\
    w/ Tea-Stu
    &  93.1\% GSR  &   84.3\% GSR    \\
    \bottomrule[1.0pt]
    \end{tabular}
    }
    \label{table: multi-skill}
\end{table}

\subsection{Unified Reward for Natural, Robust, and Generalizable Interaction Imitation}
We employ the following composite reward function across all experiments:
\begin{equation}
r_{t} = r_{t}^{\text{body}} + r_{t}^{\text{obj}} + r_{t}^{\text{rel}} + r_{t}^{c} + r_{t}^{\text{reg}},
\label{eq: imitation reward}
\end{equation}
where $r_{t}^{\text{body}}$ is the body imitation reward, $r_{t}^{\text{obj}}$ is the object imitation reward, $r_{t}^{\text{iv}}$ encourages correct relative motion between the body and the object, $r_{t}^{c}$ is the contact imitation reward, and $r_{t}^{\text{reg}}$ comprises several regularization terms to improve motion stability. Tab.~\ref{tab: obs and reward} provides a summary.

\paragraph{Body Motion Imitation Reward} 
The body reward is decomposed into multiple tracking terms:
\begin{equation}
r_{t}^{\text{body}} = r_{t}^{p} + r_{t}^{r} + r_{t}^{d} + r_{t}^{v} + r_{t}^{rv} + r_{t}^{dv} + r_{t}^{\text{amp}}.
\end{equation}
These terms correspond to body position, body rotation, joint position (DoF), body linear velocity, body angular velocity, and joint velocity, respectively. The adversarial motion prior (AMP) reward \cite{amp} is effective in enhancing the smoothness and naturalness of the body motion. Each sub-reward except $r_{t}^{\text{amp}}$ follows the general form:
\begin{equation}
r_{t}^{\alpha} = \gamma^{\alpha}\cdot\exp\left(-\lambda^{\alpha} \cdot e_{t}^{\alpha}\right),
\label{eq: rt_sub}
\end{equation}
where $e_{t}^{\alpha}$ denotes the imitation error for modality $\alpha$ at time $t$, $\lambda^{\alpha}$ and $\gamma^{\alpha}$ are hyperparameter.

\paragraph{Object Motion Imitation Reward} The object reward ensures accurate tracking of the object's state:
\begin{equation}
r_{t}^{\text{obj}} = r_{t}^{op} + r_{t}^{or},
\end{equation}
where $r_{t}^{op}$ and $r_{t}^{or}$ (optional) are the object position and rotation imitation rewards, respectively. Both are computed using the formulation in Eq.~\eqref{eq: rt_sub}. 

\paragraph{Interaction-Vector Imitation Reward} 
The IV reward $r_{t}^{\text{iv}}$ encourages correct relative interaction between robot and object. 
The IV error is computed as
\begin{equation}
e^{\text{iv}}_{t} = \bigl\| \mathbf{u}_t - \hat{\mathbf{u}}_t \bigr\|_2,
\end{equation}
where \(\hat{\mathbf{u}}_t\) is the reference IVs, and \(\mathbf{u}_t\) is the corresponding IVs from the simulation. Each vector in \(\mathbf{u}_t\) is defined as \(\mathbf{u}_t^{(k)} = \mathbf{u}_t^{k} - \mathbf{u}_t^{o}\), with \(\mathbf{u}_t^{k}\) being the 3D position of a body keypoint and \(\mathbf{u}_t^{o}\) the object position.
IV rewards deliver consistent interaction guidance even when the robot or the object deviates from the reference motion. For example, when learning from only a single ball‑pickup demonstration, we wish the robot still succeed if the ball is deviated from the reference position. In such cases, global tracking rewards like $r_{t}^{\text{body}}$
 and $r_{t}^{\text{obj}}$
 offer limited guidance, while $r_{t}^{\text{iv}}$	provides a dense and effective learning signal. This capability is crucial for generalizing interactive skills beyond the training data.

\begin{figure*}[t]
  \centering
  \vspace{-0.3cm}
  \begin{minipage}[t]{0.48\textwidth}
      \caption{Observation and Reward Settings. $\star$ denotes optional.}
        \centering
        \resizebox{\linewidth}{!}{
        \begin{tabular}{lcc|lc}
        \toprule
        \multicolumn{3}{c}{\textbf{Observations}} & \multicolumn{2}{c}{\textbf{Rewards}}\\
        \addlinespace[0.2em]
        \hline
        \addlinespace[0.25em]
        \rowcolor{gray!20} \textbf{Curr Terms} & $\pi_{\text{tea}}$ & $\pi_{\text{stu}}$ & \textbf{Mimic Terms} &
        $\pi_{\text{tea}}$ \& $\pi_{\text{stu}}$
        \\
        \midrule
        Base Ang Vel & $\checkmark$ & $\checkmark$ & Body Pos & $\checkmark$\\
        Gravity & $\checkmark$ & $\checkmark$ & Body Rot & $\checkmark$\\
        DoF Pos & $\checkmark$ & $\checkmark$ & Body Vel & $\checkmark$\\
        DoF Vel & $\checkmark$ & $\checkmark$ & Body Ang Vel & $\checkmark$\\
        Action & $\checkmark$ & $\checkmark$ & Body AMP & $\checkmark$\\
        PD Error & $\checkmark$ & $\checkmark$ & DoF Pos & $\checkmark$\\
        Ref Body Pos & $\checkmark$ &  &  DoF Vel & $\checkmark$\\
        Delta Body Pos & $\checkmark$ &  &  Interaction Vector & $\checkmark$\\
        Object Pos & $\checkmark$ & $\star$ & Object Pos & $\checkmark$\\
        Object Rot & $\star$ & $\star$ & Object Rot & $\star$\\
        Target Object Pos & $\star$ & $\star$ & Contact Graph & $\checkmark$\\
        Target Object Rot & $\star$ & $\star$ &\cellcolor{gray!20} \textbf{Reg Terms} & \cellcolor{gray!20} \\
        Skill Label &  & $\checkmark$ & Torque & $\checkmark$\\
        History & $\checkmark$ & $\checkmark$ & Action Rate & $\checkmark$\\
        \cellcolor{gray!20}\textbf{History Terms} & \cellcolor{gray!20} & \cellcolor{gray!20} & Waist DoF & $\checkmark$\\
        Base Ang Vel & $\checkmark$ & $\checkmark$ & Feet Orientation & $\checkmark$\\
        Gravity & $\checkmark$ & $\checkmark$ & Feet Slippage & $\checkmark$\\
        DoF Pos & $\checkmark$ & $\checkmark$ & DoF Limit & $\checkmark$\\
        DoF Vel & $\checkmark$ & $\checkmark$ & Torque Limit & $\checkmark$\\
        Action & $\checkmark$ & $\checkmark$ & Termination & $\checkmark$\\
        \addlinespace[0.2em]
        \hline
        \bottomrule
        \end{tabular}
        }
        \label{tab: obs and reward}
  \end{minipage}\hfill
  \begin{minipage}[t]{0.48\textwidth}
        \caption{Domain Randomization Settings.}
        \centering
        \resizebox{\linewidth}{!}{
        \label{tab: dr}
        \begin{tabular}{ll}
        \hline
        \addlinespace[0.25em]
        \rowcolor{gray!20} \textbf{Sustained External Force} &  \\
        
        Interval Range & $\mathcal{U}(1.0, 2.0)$ \\
        Body Force Magnitude Range & $\mathcal{U}(-50, 50)$ \\
        Object Force Magnitude Range & $\mathcal{U}(-2, 2)$ \\
        \rowcolor{gray!20}\multicolumn{2}{l}{\textbf{Velocity Perturbation}} \\
        Interval Range & $\mathcal{U}(1.0, 3.0)$ \\
        Velocity Magnitude Range & $\mathcal{U}(-0.5, 0.5)$ \\
        Angular Velocity Magnitude Range & $\mathcal{U}(-0.5, 0.5)$ \\
        \rowcolor{gray!20}\multicolumn{2}{l}{\textbf{Physical Properties}} \\
        DoF Friction & $\mathcal{U}(0.5, 1.25)$ \\
        Base CoM Position  & $\mathcal{U}(-0.05, 0.05)$  \\
        Default Joint DoF Offset & $\mathcal{U}(-0.01, 0.01)$ \\
        Link Mass & $\mathcal{U}(0.9, 1.1)$ \\
        P Gain & $\mathcal{U}(0.9, 1.1)$ \\
        D Gain & $\mathcal{U}(0.9, 1.1)$ \\
        Torque RFI & $0.1 \times \text{Torque Limit}$  \\
        \addlinespace[0.25em]
        \hline
        \end{tabular}
        }
  \end{minipage}
  \vspace{-0.3cm}
\end{figure*}

\paragraph{Contact Imitation Reward}
To enhance contact stability, we introduce a contact imitation reward. Our XGen pipeline annotates each frame with a contact label, indicating whether the object and key-bodies are in contact status. Inspired by prior works \cite{wang2023physhoi,wang2025skillmimic,yu2025skillmimic}, we formulate the contact state as binary vectors $\boldsymbol{s}^{cg}_t \in \{0,1\}^{J}$, where $J$ is the number of contact bodies (including objects), and a value of 1 indicates active contact.

The contact reward is computed based on the discrepancy between the simulated contact state $\boldsymbol{s}^{cg}_t$ and the reference contact state $\hat{\boldsymbol{s}}^{cg}_t$ from the reference data. The contact error vector is defined as the element-wise absolute difference:
\begin{equation}
\boldsymbol{e}^{cg}_t = |\boldsymbol{s}^{cg}_t - \hat{\boldsymbol{s}}^{cg}_t|.
\end{equation}
The contact imitation reward is then given by an exponential of the weighted error:
\begin{equation}
r_{t}^{cg} = \exp(-\sum_{j=1}^{J} \lambda^{cg}_j \cdot \boldsymbol{e}^{cg}_t[j] ),
\label{eq: rc}
\end{equation}
where $\lambda^{cg}_j$ is a sensitivity weight for the $j$-th contact edge. This formulation penalizes mismatches in the contact graph, ensuring the policy learns precise contact timing and location.

\subsection{Perceiving External Force from Proprioception}
Even in the absence of visual input, humans can still effectively perform stable grasping, basketball shooting, and other interactive behaviors. This ability relies on the implicit perception of interaction states through tactile and force feedback. In the following, we conduct a theoretical analysis to demonstrate that a similar mechanism is also feasible for humanoid robots, identify the key variables that influence such perception, and accordingly guide the observation design.

The equation of motion for a floating-base humanoid robot can be formulated as \cite{featherstone2008rigid,murray2017mathematical}:
\begin{equation}
    \boldsymbol{\tau} = \mathbf{M}(\mathbf{q})\ddot{\mathbf{q}} + \mathbf{C}(\mathbf{q}, \dot{\mathbf{q}})\dot{\mathbf{q}} + \mathbf{G}(\mathbf{q}) + \boldsymbol{\tau}_{f} + \mathbf{J}_{\text{ext}}^{\top}\mathbf{F}_{\text{ext}},
    \label{eq: equation of motion}
\end{equation}
where $\boldsymbol{\tau}$ 
denotes the vector of commanded joint torques; $\mathbf{q}$, $\dot{\mathbf{q}}$, and $\ddot{\mathbf{q}}$ are the joint positions, velocities, and accelerations, respectively; $\mathbf{M}$ is the mass matrix; $\mathbf{C}$ captures Coriolis and centrifugal terms; $\mathbf{G}$ is the gravity vector; $\boldsymbol{\tau}_{f}$ accounts for joint friction; and $\mathbf{J}_{\text{ext}}^{\top}\mathbf{F}_{\text{ext}}$ represents the joint-space projection of external contact forces $\mathbf{F}_{\text{ext}}$.
Rearranging Eq.~\ref{eq: equation of motion} yields
\begin{equation}
\mathbf{J}_{\text{ext}}^{\top}\mathbf{F}_{\text{ext}} = \boldsymbol{\tau} - \bigl(\mathbf{M}\ddot{\mathbf{q}} + \mathbf{C}\dot{\mathbf{q}} + \mathbf{G} + \boldsymbol{\tau}_{f}\bigr),
\label{eq: force}
\end{equation}
which shows that the external forces on each joint can be estimated if corresponding parameters are available. In the context of training a whole‑body control policy with RL, this implies that, in principle, a policy can possibly learn to implicitly perceive external forces—provided it is given access to the relevant observations—and thus achieve a better dynamic interaction with objects. Taking the Unitree G1 as an example, $\mathbf{q}$ and $\dot{\mathbf{q}}$ are directly measurable, and $\boldsymbol{\tau}$ is well approximated by the commanded torque $\boldsymbol{\tau}_{\text{cmd}}$ from the PD controller. Although the joint acceleration $\ddot{\mathbf{q}}$ is not directly available, we can implicitly provide coarse acceleration information by including a history of multiple frames of $\dot{\mathbf{q}}$ in the observations. The remaining terms in Eq.~\ref{eq: force} are approximately constant and therefore do not need to be provided to the policy. 

Based on these insights, the final observation components used in our policy are listed in Tab.~\ref{tab: obs and reward}; they account (either explicitly or implicitly) for all variables appearing in Eq.~\ref{eq: force}.

\subsection{Domain Randomization}
As summarized in Fig.~\ref{tab: dr}, we apply domain randomization (DR) to various physical properties \cite{tobin2017domain}, including object size, mass, and coefficient of restitution, as well as robot friction coefficients, center of mass offsets, and perception noise. Additionally, we apply sustained random external forces to the robot during training. These DR terms are particularly important for achieving robust deployment.

\subsection{Probabilistic Early Termination}

To improve training efficiency, we employ early termination~\cite{peng2018deepmimic} when the simulation enters an undesirable state. In addition to the \textbf{Interaction Termination (IT)} described in the main text, we consider the following types of termination: Root Orientation Tracking Termination, Body Tracking Termination, Object Tracking Termination, and Motion End Termination.

Similar to IT, all the above terminations (except for Motion End Termination) are implemented \emph{probabilistically}. This design ensures that the policy can, with a certain probability, continue training even when a termination condition is met. The purpose is to enhance deployment stability: because real‑world execution cannot be terminated as in simulation, the agent may encounter states that would otherwise trigger termination. Allowing the policy to occasionally experience and learn from such states improves its robustness when deployed on physical hardware.

\subsection{PD Controller}
The low-level joint motor control is realized by a PD controller that computes the desired joint torque $\tau_j$ for each joint $j$ at each time step, formulated as:
\begin{equation}
\tau_j = k_{p,j} \; (q_j^{\text{ref}} - q_j) \; - \; k_{d,j} \; \dot{q}_j,
\end{equation}
where $q_j^{\text{ref}}$ and $q_j$ are the target and measured joint positions for joint $j$, respectively, and $\dot{q}_j$ is the measured joint velocity. The proportional and derivative gains, $k_{p,j}$ and $k_{d,j}$, are derived following previous work~\cite{liao2025beyondmimic}, formulated as:
\begin{equation}
k_{p,j} = I_j \omega^2, \quad k_{d,j} = 2 I_j \zeta \omega
\end{equation}
where \(\omega\) denotes the natural frequency, \(\zeta\) is the damping ratio, and \(I_j = k_{g,j}^2 I_{\text{rotor},j}\) represents the reflected inertia of the joint. Here, \(I_{\text{rotor},j}\) is the rotor inertia and \(k_{g,j}\) is the gear ratio. We set \(\zeta = 2\) and \(\omega = 10 \, \text{Hz}\).

The PD controller's target position \(q_j^{\text{ref}}\) is formulated as:
\begin{equation}
q_j^{\text{ref}} = q_j^0 + \alpha \cdot a_{j,t},
\end{equation}
where \(q_j^0\) and \(a_{j,t}\) represents the joint position offset and policy action for joint $j$. The action scale \(\alpha\) is defined as:
\begin{equation}
\alpha = 0.25 \cdot \frac{\tau_{j,\text{max}}}{k_{p,j}}
\end{equation}
where \(\tau_{j,\text{max}}\) is the torque limit for joint \(j\).

\subsection{Simulating MoCap Signal Loss}
\label{app:packet_loss}

In real-world motion capture systems, tracking object states during interactive tasks frequently suffers from packet loss. This loss manifests primarily in two modes: 
\begin{itemize}
\item A: the signal loss occur intermittently even under good exposure and favorable conditions
\item B: the signal loss becomes severe as the object moves close to the robot (e.g., near the hands or torso), significantly raising the missing rate.
\end{itemize}

To improve deployment robustness, we simulate both packet‑loss modes during training. At the start of each episode, a distance threshold is sampled uniformly from a predefined range. During execution, the distance between the object and the anchor (e.g., the robot’s hand center) is computed at every simulation step. If this distance falls below the sampled threshold, the episode operates under B mode; otherwise, it follows A mode.

Each mode is parameterized by a distinct packet‑loss probability and loss‑duration window (i.e., the maximum number of consecutive frames that can be lost). Mode A uses a relatively low loss probability and a short loss window, while mode B adopts substantially higher values for both parameters. When a loss event is active, the object‑related observation vector is masked (set to zero); other observations remain unchanged. Parameters are manually tuned so that the statistics of the simulated object‑perception signal approximate those of a physical MoCap system in real operation.

\end{document}